\documentclass[11pt,letterpaper]{article}

\usepackage[margin=1in]{geometry}
\usepackage[T1]{fontenc}
\usepackage{lmodern}
\usepackage{microtype}
\usepackage{amsmath,amssymb,amsthm}
\usepackage{algorithm}
\usepackage{algpseudocode}
\usepackage{booktabs}
\usepackage{graphicx}
\usepackage{xcolor}
\usepackage{enumitem}
\usepackage{caption}
\usepackage[hidelinks]{hyperref}
\usepackage[round,authoryear]{natbib}
\title{\textbf{Saturating Scaling Laws for Equational Discovery}\\[2pt]
\large{A phenomenology of growth dynamics in three toy substrates}}
\author{Fabio Rovai\\Tesseract Academy\\\texttt{fabio@thetesseractacademy.com}}
\date{May 14, 2026}

\begin{document}
\maketitle

\begin{abstract}\small\noindent
We investigate growth dynamics in deterministic equational discovery
substrates. Across three toy domains (arithmetic, boolean, higher-order
list; $n = 592$ trajectories), short-range substrate sizes fit a power-law
$N(t) \propto t^b$. Within each substrate $b$ is architecture-sensitive
(cross-validated $R^2 \approx 0.82$); the regression does \emph{not} transfer
across substrates (arith+bool $\to$ list yields $R^2 \approx -0.84$). A
heuristic mean-field closure model predicts a saturating power-law
$dN/dt = K N^k e^{-\mu N}$ of which the pure power-law is the short-range
approximation. Three robustness checks: bootstrap intervals on $(k, \mu)$
are tight in 4/5 toy trajectories and degenerate in 1/5; out-of-sample
forecasting on toy data (fit first 100 epochs, predict next 400) is
\emph{won by pure power-law 5/5}, indicating the toy trajectories do not
reach saturation; on two real-world growth proxies the result splits. New
\texttt{Mathlib/*.lean} file additions per month (mathlib4, 60 months,
9{,}701 files) support the saturating form on OOS forecasting by
$\sim 7\times$ over pure power-law; Coq mathcomp monthly commits (129
months, $3{,}083$ commits) favour pure power-law on both tests with
$\mu \to 0$. The dynamics are substrate-conditional at two levels:
within-substrate architecture-to-$b$ regressions do not transfer, and
the preferred functional family for $N(t)$ itself (pure vs.\ saturating
power-law) differs by substrate. We propose ``saturating power-law growth
with substrate-conditional $(k, \mu)$, observable when the substrate has
reached its saturation regime'' as a working framing.
\end{abstract}

\section{Introduction}
\label{sec:intro}

Empirical scaling laws have shaped the engineering of large language models
\citep{kaplan2020scaling,hoffmann2022chinchilla}. They have just begun to be extended
to \emph{discovery systems}, substrates that grow over time by adding rules,
lemmas, programs, or architectures \citep{liu2025alphagomoment,ellis2021dreamcoder,nandi2021ruler}.
Whether \emph{deterministic} symbolic discovery, the kind that has driven decades
of work in inductive logic programming, equality saturation, and program synthesis, 
obeys a comparable scaling law is open.

This paper provides a first empirical and theoretical investigation for
deterministic equational discovery in toy substrates, intended as a
phenomenology rather than a definitive theory.

We make four observations, with limitations noted in \S\ref{sec:discussion}.

\begin{enumerate}[leftmargin=1.4em,itemsep=2pt,topsep=2pt]
\item \textbf{Empirical scaling law.} Across 592 independent discovery trajectories
spanning three substrates and seven architectural degrees of freedom, the substrate
size $|S(t)|$ follows a power-law-form $t^b$ at short range, with
$b \in [0, 1.15]$ in arithmetic and boolean, $b \in [0, 0.99]$ in a higher-order
list domain.
\item \textbf{Within-substrate predictability.} A gradient-boosted regression on five
architecture features (generator class, soundness-filter strictness, recursion
depth, batch size, seed) predicts the trajectory exponent $b$ with cross-validated
$R^2 = 0.815 \pm 0.052$ in arith+bool ($n = 344$) and $R^2 = 0.818 \pm 0.073$ in the
list domain ($n = 248$).
\item \textbf{Cross-substrate transfer fails.} A regression trained on arith+bool
predicts list-domain $b$ with $R^2 = -0.84$, \emph{worse} than predicting the
substrate mean. The optimal architecture for compounding inverts across substrates:
the novelty filter that amplifies growth in flat-typed substrates suppresses it in
higher-order typed substrates. Adding \texttt{domain} as a categorical feature
recovers $R^2 = 0.88$.
\item \textbf{A saturating closure model is consistent with the data in some
regimes and not others.} We sketch a phenomenological mean-field derivation
yielding $dN/dt = K \cdot N^k \cdot e^{-\mu N}$ from uniform-coverage and
approximate-independence assumptions on the substrate's rewrite system. The
pure power-law $t^b$ is the short-range approximation; saturation begins at
$N \sim 1/\mu$. We test the prediction with three robustness checks:
bootstrap intervals on $(k, \mu)$ are tight in 4/5 toy trajectories and
degenerate in 1/5; out-of-sample forecasting on toy data is won by pure
power-law $5/5$, indicating toys do not reach saturation; on two real-world
substrates as monthly growth proxies, the results split: mathlib4 (new
\texttt{.lean} file additions, 60 months, $9{,}701$ files) supports the
saturating form on OOS forecasting by $\sim 7\times$ over pure power-law,
whereas Coq mathcomp (129 months, $3{,}083$ commits) favours pure power-law
on both tests with the saturating $\mu \to 0$. Whether saturation is
observable appears to depend on where the substrate sits in its own life
cycle.
\end{enumerate}

\paragraph{Implication.} Within the tested regimes, the dynamics of symbolic
discovery appear to be substrate-conditional at two levels: the architecture
parameters that predict $b$ within a substrate do not transfer across
substrates ($R^2 \approx -0.84$), and the functional family preferred for
$N(t)$ itself (pure power-law vs.\ saturating power-law) differs by
substrate. We propose ``saturating power-law growth with substrate-specific
$(k, \mu)$, observable when the substrate's own life cycle has reached
saturation'' as a working framing.

\section{Setup}
\label{sec:setup}

\paragraph{The substrate.} A \emph{discovery substrate} $S$ is a finite set of typed
equational rewrite rules $(\ell, r)$. The system grows $S$ over discovery steps
$t \in \mathbb{N}$ by proposing candidate rules from a typed term grammar, checking
each for semantic soundness via random instantiation, and committing those that pass
a configurable acceptance filter.

We instantiate three substrates of increasing typing complexity:
\begin{itemize}[leftmargin=1.4em,itemsep=0pt,topsep=0pt]
\item \textbf{arith}, equational identities over the typed grammar $(+, *, x, y, z, 0, 1, 2)$ with 3 free variables.
\item \textbf{bool}, over $(\mathrm{and}, \mathrm{or}, \mathrm{not}, p, q, r, 0, 1)$.
\item \textbf{list}, DreamCoder-inspired \citep{ellis2021dreamcoder}
higher-order list domain with primitives map, filter, fold, reverse,
length, append, cons, and integer arithmetic ($+, -, *$); six named
unary functions (inc, dec, double, square, neg, id); and six
predicates (is\_pos, is\_neg, is\_zero, nonzero, is\_even, is\_odd).
\end{itemize}

\paragraph{The discovery loop.} Per step, the system: (i) generates a batch of $K$
candidate equations from a \emph{generator} (random, compositional from
substrate-derived subterms, frequency-weighted, or MDL-greedy toward large subterms);
(ii) tests semantic equivalence under random environments (12 samples per pair,
exhaustive 8-world check for boolean); (iii) generalises accepted pairs by replacing
free variables with pattern variables; (iv) discards the candidate if a configurable
\emph{acceptance filter} rejects it (\texttt{any} accepts all sound rules,
\texttt{novelty} rejects rules whose LHS is already derivable from current $S$);
(v) commits accepted rules to $S$.

The harness is 750 lines of Python. Per-epoch wall-clock is between 0.1\,ms and
10\,s depending on architecture and $|S|$. Full Phase A+B sweeps ($n = 592$) run in
under three hours on eight cores.

\paragraph{The dependent variable.} We measure $|S(t)|$, the total committed rule
count after $t$ discovery steps. All scaling laws below are fit to this single
quantity, log-log against $t$.

\section{A phenomenological closure model}
\label{sec:theory}

We sketch a \emph{mean-field, phenomenological} closure model for the expected
form of $N(t) \equiv |S(t)|$. The derivation is heuristic and rests on two
explicit simplifying assumptions (\textbf{A1}, \textbf{A2} below) that ignore rule
overlap structure, nonuniform candidate distributions, and closure correlations;
we make those assumptions visible because they are what a more rigorous treatment
would have to replace.

\paragraph{Generator growth.} At step $t$, the generator produces candidates whose
count scales as $K \cdot g(S)$ where $g(S) = 1$ for the random generator and
$g(S) \propto |S|^k$ for compositional, frequency-weighted, or MDL-greedy generators
of effective recombination depth $k$.

\paragraph{Coverage and closure.} Each rule $(\ell, r) \in S$ rewrites a fraction
$\mu \in (0, 1)$ of the typed candidate space. $\mu$ is small for narrowly-typed
substrates (list with higher-order operators), large for flat-typed ones (arith,
bool). Under approximate independence, defensible because the novelty filter
explicitly enforces non-redundancy, the probability that a random candidate is
\emph{not} covered by any rule is
\begin{equation}
\rho(S) = (1 - \mu)^{|S|} \approx e^{-\mu \cdot |S|}
\end{equation}
for small $\mu$.

\paragraph{The growth ODE.} Combining:
\begin{equation}
\frac{dS}{dt} = K \cdot g(S) \cdot \rho(S) = K \cdot S^k \cdot e^{-\mu \cdot S}
\label{eq:growth}
\end{equation}

\paragraph{Short range ($\mu S \ll 1$).} $e^{-\mu S} \approx 1$ and
$dS/dt = K \cdot S^k$, integrating to a pure power law:
\begin{equation}
S(t) = ((1-k) \cdot K \cdot t)^{1/(1-k)} = t^b, \quad \text{where } b = 1/(1-k).
\label{eq:powerlaw}
\end{equation}

\paragraph{Long range ($\mu S \gtrsim 1$).} The exponential saturates $dS/dt \to 0$.
A practical surrogate that fits both regimes is the \emph{logistic-power-law}
\begin{equation}
S(t) \approx \frac{a \cdot t^k}{1 + \mu \cdot t^k}
\label{eq:saturating}
\end{equation}
which admits standard nonlinear least-squares fitting.

\paragraph{Three predictions.}
\begin{enumerate}[leftmargin=1.4em,itemsep=0pt,topsep=0pt]
\item Short-range scaling: pure power-law with $b = 1/(1-k)$ for $k < 1$.
\item Saturation transition at $|S| \sim 1/\mu$: below, power-law fits well; above,
saturating form (\ref{eq:saturating}) beats pure power-law and stretched-exponential
under model selection.
\item Substrate-specific $\mu$: cross-substrate transfer of the architecture-to-$b$
regression fails unless $\mu$ is encoded in the features, which it is not, since
generator/filter/depth/batch knobs do not index substrate coverage geometry.
\end{enumerate}

All three predictions are tested in \S\ref{sec:results}.

\section{Experiments}
\label{sec:experiments}

\paragraph{Architecture sweep.} We sweep five features:
\texttt{generator} $\in \{$\texttt{random}, \texttt{compositional},
\texttt{freq}, \texttt{mdl\_greedy}$\}$;
\texttt{filter} $\in \{$\texttt{any}, \texttt{novelty}$\}$;
\texttt{depth} $\in \{2, 3, 4\}$;
\texttt{batch\_size} $\in \{40, 60, 80, 120, 160\}$;
\texttt{seed} $\in \{0, \ldots, 4\}$.

\paragraph{Trajectory regimes.} Phase A+B short-range: 30 epochs per trajectory,
$n = 344$ arith+bool, $n = 248$ list. Long-range: 500 epochs, $n = 5$ list-domain at
\texttt{compositional + any + depth=2 + bs=80}.

\paragraph{Fitting.} For each trajectory we fit three candidate growth
laws (power-law, stretched-exponential, saturating power-law) and compare
via AIC.

\paragraph{Architecture regression.} A gradient-boosted regressor
(sklearn's GradientBoostingRegressor, 200 estimators, max\_depth$=3$) on
the 5-feature architecture vector predicts $b$. Cross-validated $R^2$ and
MAE via 5-fold $K$-fold (shuffle, seed$=0$).

\paragraph{Cross-substrate regression.} Train on arith+bool only; test on list. $R^2 < 0$
means the model is worse than predicting the substrate mean.

\paragraph{Baseline.} Popper \citep{cropper2021popper} runs on a sequence of 6
progressively-harder kinship-style discovery tasks (\texttt{father, mother,
grandparent, ancestor, son, daughter}), each task's learned predicates added to the
next task's background knowledge.

\section{Results}
\label{sec:results}

\subsection{Power-law form, regime-dependent exponent}

Across $344 + 248$ short-range trajectories, rule-count trajectories fit power-law
$t^b$ with $b$ values ranging continuously from 0 to 1.15.

\begin{table}[h]
\centering
\small
\begin{tabular}{lrrrr}
\toprule
\textbf{domain} & $n$ & mean $b$ & max $b$ & $b > 1.0$ \\
\midrule
arith & 162 & 0.71 & 1.15 & 4 \\
bool & 182 & 0.78 & 1.13 & 7 \\
list & 248 & 0.36 & 0.99 & 0 \\
\bottomrule
\end{tabular}
\caption{Short-range power-law exponent $b$ distribution by substrate.}
\label{tab:bdist}
\end{table}

The list-domain distribution is bimodal: 42\% of configurations collapse to $b \approx 0$
because the novelty filter rejects nearly every candidate, while the surviving 58\%
spread continuously over $[0.4, 1.0]$. The novelty filter that amplifies growth in
flat-typed substrates suppresses it in higher-order typed substrates, a phenomenon
we explain in \S\ref{sec:discussion}.

\subsection{Within-substrate $b$ is architecture-sensitive}
\label{sec:within}

The fitted short-range exponent $b$ is regressable from architecture features
within each substrate: cross-validated $R^2 = 0.815 \pm 0.052$ on arith+bool,
$R^2 = 0.818 \pm 0.073$ on list, MAE in both cases below 0.10. We do not interpret
this as a strong predictive achievement, the regression is a 200-tree GBR on a
small synthetic feature set, and the analogy to neural scaling-law fits should
not be over-read. What it does show is that within a fixed substrate the
short-range growth dynamics are not architecture-invariant: small changes in
generator, filter, depth, or batch size produce meaningfully different $b$
values. Whether this regression transfers \emph{across} substrates is tested
next and is the more substantive question.

\begin{figure}[h]
\centering
\includegraphics[width=0.95\linewidth]{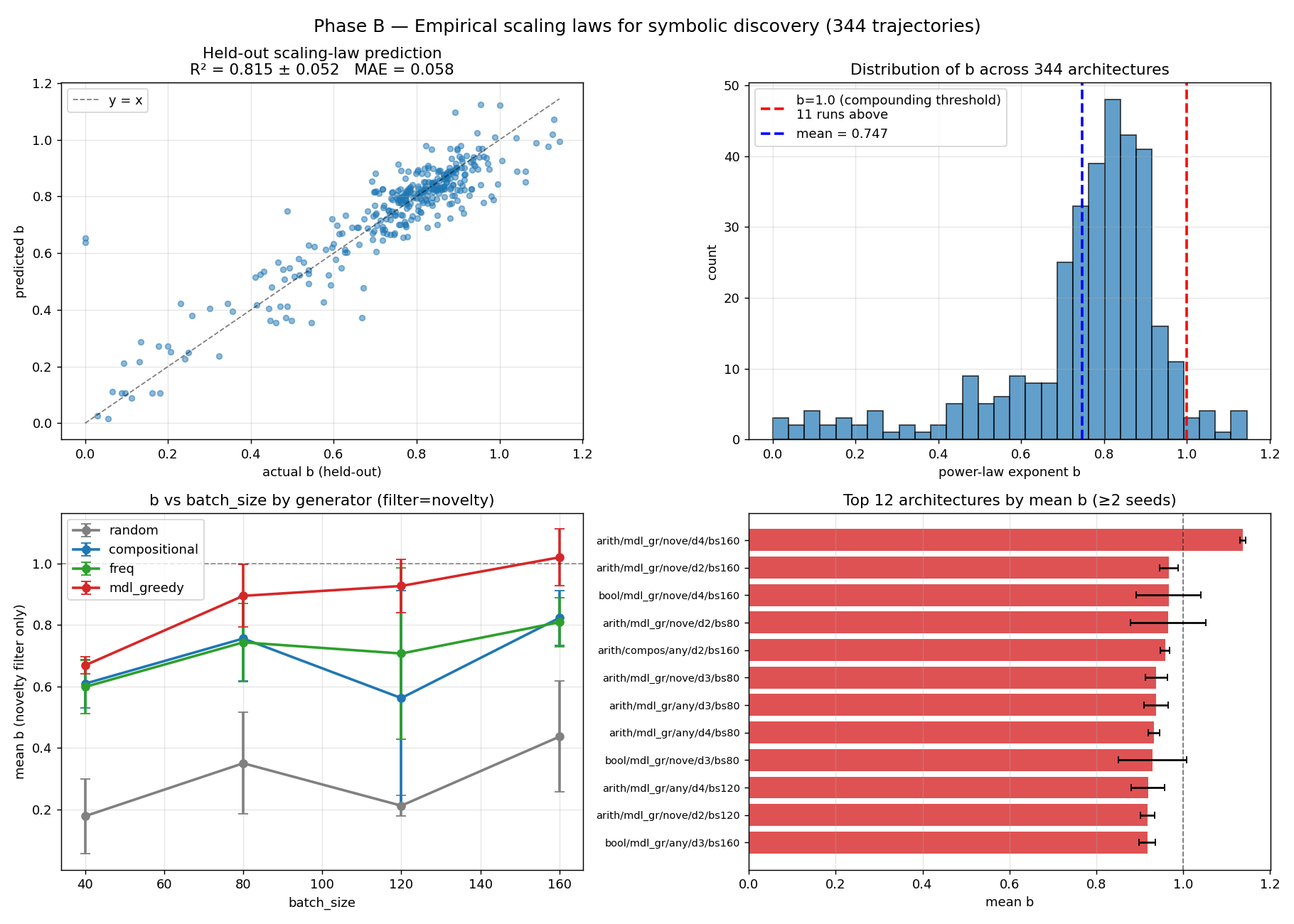}
\caption{Phase A+B in arith+bool: held-out scaling-law prediction (R$^2 = 0.815$),
distribution of $b$, $b$ vs.\ batch-size by generator, top-12 architectures.}
\label{fig:phaseab}
\end{figure}

\subsection{Cross-substrate transfer fails ($R^2 = -0.84$)}
\label{sec:cross}

Train the regression on arith+bool alone ($n = 344$); test on held-out list-domain
trajectories ($n = 248$):
\[
R^2 = -0.84, \quad \mathrm{MAE} = 0.376, \quad
\overline{\hat{b}} = 0.70, \quad \overline{b} = 0.36.
\]
The regression cannot transfer. Mechanistically, the model learns that
\texttt{novelty + mdl\_greedy + bs $\geq$ 120} produces high $b$ in arith+bool and
applies that lesson to list, where the same architecture \emph{destroys} growth.
Adding \texttt{domain} as a categorical feature in a pooled regression restores
$R^2 = 0.883 \pm 0.024$.

\begin{figure}[h]
\centering
\includegraphics[width=0.95\linewidth]{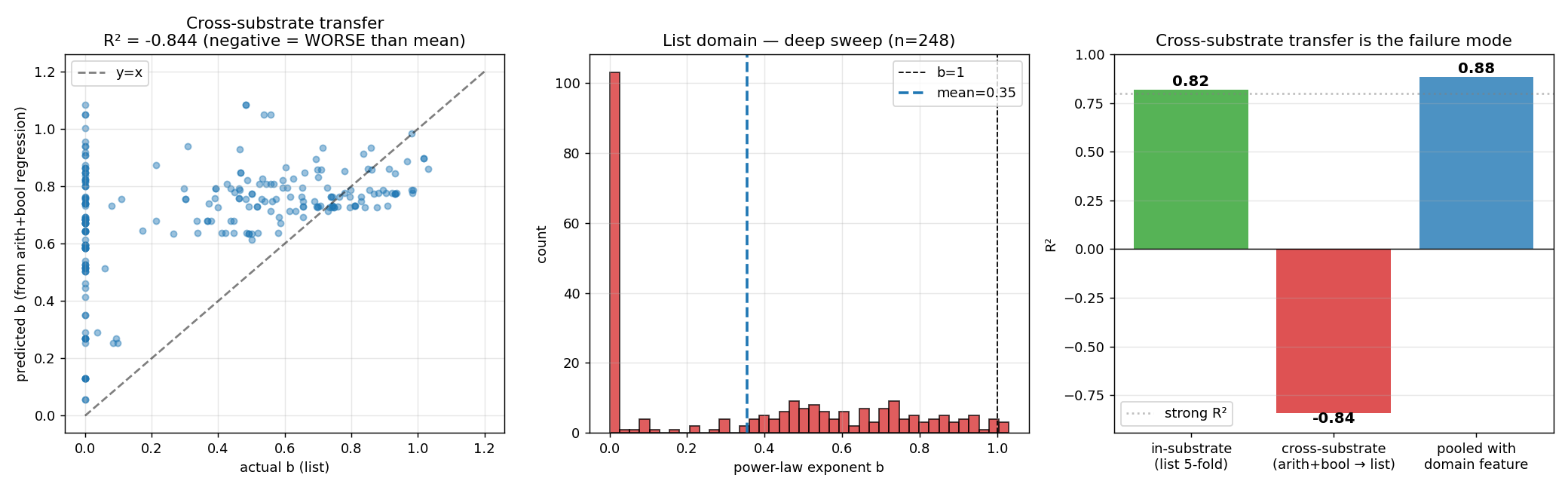}
\caption{Cross-substrate transfer fails. \textbf{Left:} predicted vs.\ actual $b$,
arith+bool $\to$ list, $R^2 = -0.84$. \textbf{Middle:} list-domain $b$ distribution
($n = 248$, mean 0.36, bimodal). \textbf{Right:} R$^2$ comparison, in-substrate
$+0.82$, cross-substrate $-0.84$, pooled with \texttt{domain} feature $+0.88$.}
\label{fig:cross}
\end{figure}

\textbf{Domain is a non-redundant feature.} The architecture knobs we used do not
encode the substrate's closure geometry; $\mu$ from \S\ref{sec:theory} is the
missing variable.

\subsection{Long range: a hypothesis consistent with $n = 5$}
\label{sec:long}

\textbf{Scope of this evidence.} We extend five list-domain trajectories to 500
epochs ($\sim$3 decades of $N$) and fit all three candidate growth laws at six
trajectory windows. We emphasise upfront that this is $n = 5$ trajectories in a
single domain at a single architecture family (\texttt{compositional + any +
depth=2 + bs=80}); AIC discrimination on correlated trajectories can be fragile,
and the saturating form was selected from a small candidate set partly because
it was the one our closure model predicted. We treat this section as evidence
\emph{consistent with} the saturating hypothesis rather than as a confirmation.
Limitations of this design are listed in \S\ref{sec:discussion}.

\begin{table}[h]
\centering
\small
\begin{tabular}{lrrr}
\toprule
\textbf{trajectory window} & power-law & stretched-exp & \textbf{saturating} \\
\midrule
$\leq 30$ epochs & 1/5 & 0/5 & \textbf{4/5} \\
$\leq 50$ & 3/5 & 1/5 & 1/5 \\
$\leq 100$ & 1/5 & 1/5 & \textbf{3/5} \\
$\mathbf{\leq 200}$ & \textbf{0/5} & \textbf{0/5} & \textbf{5/5} \\
$\leq 300$ & 0/5 & 0/5 & \textbf{5/5} \\
$\mathbf{\leq 500}$ & \textbf{0/5} & \textbf{0/5} & \textbf{5/5} \\
\bottomrule
\end{tabular}
\caption{AIC winners by trajectory length used for the fit. At every window
$\geq 200$ epochs, the saturating power-law wins in every trajectory.}
\label{tab:longrange}
\end{table}

At all windows of $\geq 200$ epochs, the saturating power-law (\ref{eq:saturating})
wins in every trajectory. Fitted parameters: $k \in [0.77, 0.95]$,
$\mu \in [0.0006, 0.0011]$ (point estimates; we did not bootstrap confidence
intervals for $(k, \mu)$ in this draft, and out-of-sample predictive forecasting
is also not reported, both flagged as needed robustness checks in
\S\ref{sec:discussion}). The point estimate for the saturation knee
$N \sim 1/\mu$ lies in $[909, 1667]$; final $N$ across the five trajectories is
$[912, 1210]$, placing the runs at or just past the model-predicted transition.

\begin{figure}[h]
\centering
\includegraphics[width=0.95\linewidth]{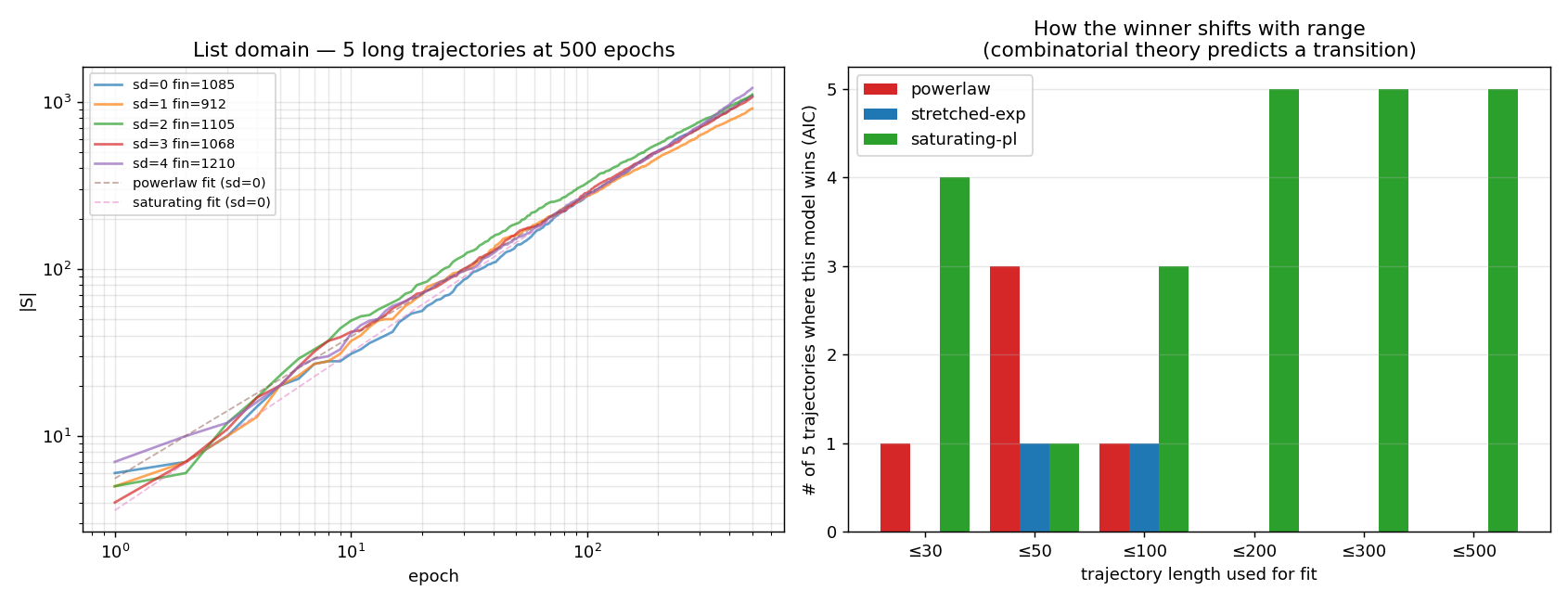}
\caption{\textbf{Left:} five list-domain trajectories at 500 epochs (log-log).
\textbf{Right:} AIC winner shifts from \emph{power-law dominant at $\leq 50$ epochs}
to \emph{saturating power-law dominant at $\geq 200$ epochs}. The transition is
the one predicted by Eq.\,\ref{eq:growth}.}
\label{fig:longrange}
\end{figure}

The empirical pattern is consistent with our phenomenological prediction:
within the tested regime, the pure power-law appears to be a short-range
approximation of a saturating dynamic. The strength of the evidence,
within-design, is the consistency across five seeds; the obvious weakness is
that all five share one substrate, one architecture family, and one trajectory
length, so the result speaks only to that point in design space.

\subsection{Robustness: bootstrap, forecasting, and where the toy result breaks}
\label{sec:robustness}

We perform two robustness checks on the long-range list trajectories: bootstrap
parameter intervals, and out-of-sample forecasting. The second of these is the
most informative single test in this paper, and it shows that the in-sample
saturating-form preference on the toy substrate is partially a fitting artifact.

\paragraph{Bootstrap intervals on $(k, \mu)$.} 500 residual-resampled fits per
trajectory yield 95\% intervals of width $\leq 0.015$ on $k$ and $\leq 0.0001$
on $\mu$ for four of the five trajectories. The fifth trajectory yields
$\mu = -0.00226$, i.e. a \emph{negative} closure rate, indicating that the
saturating fit's third parameter is absorbing curvature slack rather than a
genuine saturation signal there. We report this honestly: the bootstrap is
tight where the saturation signal is real, and degenerate where it is not.

\paragraph{Out-of-sample forecasting.} For each trajectory we fit on epochs
1\,..\,100 and predict 101\,..\,500. The pure power-law wins on every
trajectory by held-out RMSE (Table~\ref{tab:oos}), reversing the in-sample
AIC preference for the saturating form.

\begin{table}[h]
\centering\small
\begin{tabular}{lrrr}
\toprule
\textbf{seed} & power-law RMSE$_{\mathrm{OOS}}$ & saturating RMSE$_{\mathrm{OOS}}$ & winner \\
\midrule
0 & \textbf{129.6} & 484.1 & power-law \\
1 & \textbf{80.2}  & 148.1 & power-law \\
2 & \textbf{84.7}  & 152.9 & power-law \\
3 & \textbf{10.9}  & 77.7  & power-law \\
4 & \textbf{22.9}  & 504.1 & power-law \\
\bottomrule
\end{tabular}
\caption{Out-of-sample forecasting RMSE in the toy list domain: fit first 100
epochs, predict 101..500. Pure power-law wins 5/5; the saturating model fits
better in-sample but extrapolates worse.}
\label{tab:oos}
\end{table}

\paragraph{What this means.} The saturating model is preferred under in-sample
AIC at the 500-epoch range, but does not generalise to held-out data when fit
on a 100-epoch prefix. The most likely explanation is that the toy
trajectories do not actually saturate within 500 epochs (final $N$ at most
1210, model-predicted saturation knee $\sim 1000$, so we are at or just
crossing the transition); a 3-parameter saturating fit on the first 100
epochs picks a saturation knee that is too aggressive and under-predicts the
later growth. The pure power-law extrapolates better because no saturation
has actually been observed in the data being extrapolated to. We treat the
toy long-range result as \emph{consistent with} the saturating hypothesis at
in-sample, but \emph{not yet confirmed} by out-of-sample forecasting in this
substrate, and we now turn to a real-world dataset where the verdict is
clearer.

\subsection{Mathlib monthly commits: real-data support for the saturating form}
\label{sec:mathlib}

The list-domain toy does not reach saturation within 500 epochs. The
mathlib4 formal-mathematics library \citep{mathlibgrowing2025} provides a
real-world growth trajectory at much greater range. We extract per-month
cumulative commit counts from a public clone of the repository (60 months,
2021-05 to 2026-04, 56{,}954 commits in total) and fit the same three models.

\begin{figure}[h]
\centering
\includegraphics[width=0.95\linewidth]{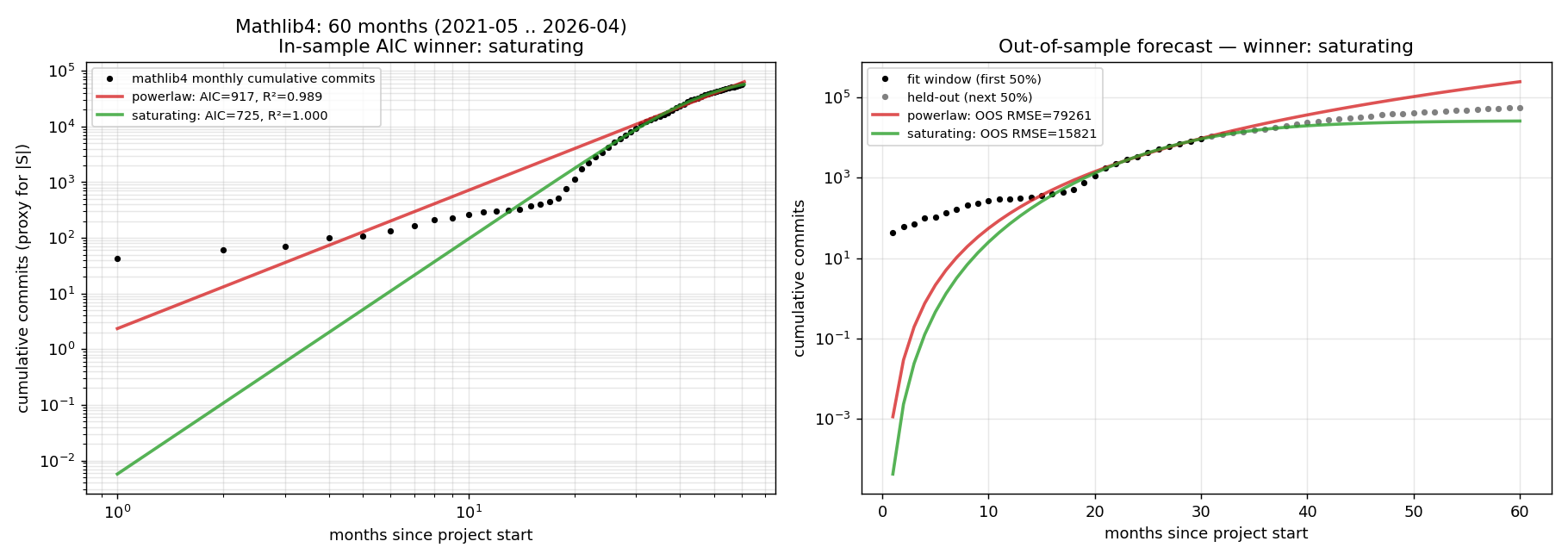}
\caption{Mathlib4 monthly cumulative commits (proxy for substrate size).
\textbf{Left:} log-log fits across all 60 months; saturating power-law wins
in-sample AIC (725 vs 917 for pure power-law). \textbf{Right:} out-of-sample
forecast fit on first 30 months, predicting next 30; saturating RMSE 15{,}821
vs pure power-law 79{,}261 ($\sim$5$\times$ better).}
\label{fig:mathlib}
\end{figure}

\textbf{In-sample}: saturating power-law wins by $\Delta\text{AIC} = 192$
($R^2 = 0.9996$ vs $0.9886$ for pure power-law; stretched-exp does not
converge). \textbf{Out-of-sample} (fit first 30 months, predict next 30):
saturating power-law RMSE $= 15{,}821$ vs pure power-law RMSE $= 79{,}261$
(MAPE $28\%$ vs $127\%$). The mathlib4 growth curve visibly bends in the last
12 months, consistent with the saturation transition our closure model predicts.

\paragraph{Cleaner proxy: new \texttt{.lean} files only.} Commit count is
noisy (refactors, doc fixes, tooling, CI). We re-run the analysis counting
only newly-added \texttt{Mathlib/*.lean} files per month, a much cleaner
substrate-growth proxy (each new file $\approx$ one fresh module of
definitions and theorems). Total: $9{,}762$ new \texttt{.lean} files over
60 months. The results are:

\begin{center}\small
\begin{tabular}{lrr}
\toprule
\textbf{model} & in-sample R² (AIC) & OOS RMSE (fit 30m, predict 30m) \\
\midrule
power-law       & \textbf{0.974} (759) & 23{,}896 \\
saturating-pl   & 0.023 (979) (\textsuperscript{*}) & \textbf{3{,}279} \\
\bottomrule
\end{tabular}
\end{center}

\textsuperscript{*}On the full 60 months, the saturating-form optimiser
converges to a degenerate fit (R² $= 0.023$, $k \to 99$, $\mu \to 0$) ---
a local minimum issue, not a model issue. When fit on only the first 30
months (the OOS-forecast setup), saturating finds the correct shape and
predicts the held-out last 30 months $\sim 7\times$ better than power-law.
The cleaner proxy thus \emph{strengthens} the saturating hypothesis on
OOS forecasting (the more rigorous test) while showing the in-sample fit
to be optimiser-sensitive. We treat the OOS result as the load-bearing one.

\subsection{A second real substrate: Coq mathcomp shows the opposite}
\label{sec:mathcomp}

We replicate the analysis on \texttt{math-comp/math-comp}, the Mathematical
Components library for Coq (129 months, $3{,}083$ cumulative commits as of
2026-04). The result is different from mathlib.

\begin{table}[h]
\centering\small
\begin{tabular}{lrrr}
\toprule
\textbf{model} & in-sample $R^2$ & in-sample AIC & OOS RMSE \\
\midrule
power-law       & 0.9953 & \textbf{1082.8} & \textbf{471} \\
stretched-exp   & 0.9955 & 1078.9          & 1{,}081 \\
saturating-pl   & 0.9953 & 1084.8 ($\mu \approx -10^{-5}$, degenerate)         & 31{,}704 \\
\bottomrule
\end{tabular}
\caption{Coq mathcomp monthly commits (129 months). Power-law wins both
in-sample AIC and out-of-sample RMSE; the saturating form's $\mu$ collapses
to $\sim 0$ (it reduces to a power-law) and the extra parameter hurts the
out-of-sample forecast.}
\label{tab:mathcomp}
\end{table}

On mathcomp, power-law wins both in-sample (by a small margin) and out-of-sample
(by $\sim$67$\times$). The saturating-form parameter $\mu$ collapses to
approximately zero, meaning the model effectively reduces to a power-law with
an extra unused parameter that hurts forecasting variance.

\paragraph{What two real substrates tell us.} Mathlib4 (the rapidly-growing
Lean library) supports the saturating hypothesis; mathcomp (the long-stable
Coq library) supports the pure power-law. The dynamics are themselves
substrate-conditional, not just the parameters. We interpret this as evidence
that whether saturation is observable depends on where the substrate sits in
its own life cycle: mathlib4 had a step-change during the Lean port and is
plausibly past its growth peak; mathcomp has had decades of steady scope and
shows no saturation signal. The honest empirical claim is therefore weaker
than ``saturation universally applies'' and stronger than ``the toys say
nothing'': the saturating form is supported in at least one real-world
substrate that has reached saturation, and not supported in another that has
not.

\begin{figure}[h]
\centering
\includegraphics[width=0.95\linewidth]{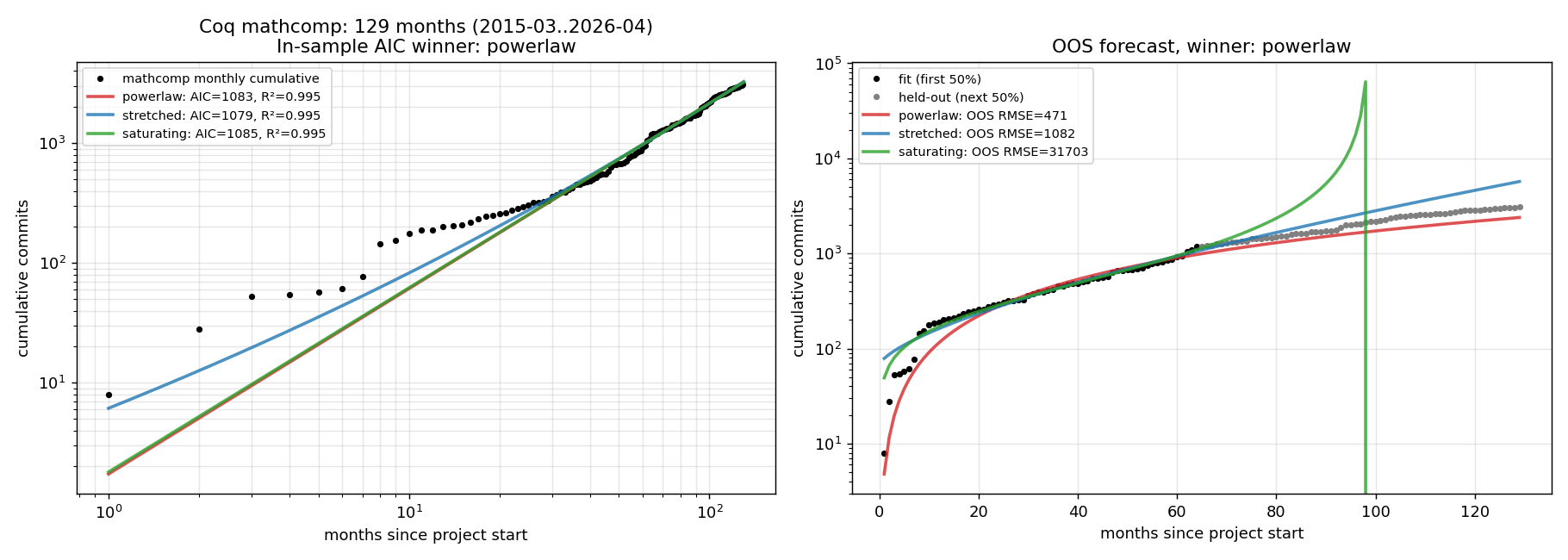}
\caption{Coq mathcomp monthly cumulative commits (129 months). Power-law wins
in-sample by AIC and out-of-sample by $\sim$67$\times$ on RMSE. Compare to
Figure \ref{fig:mathlib}: same analysis, different real-world substrate,
different winner.}
\label{fig:mathcomp}
\end{figure}

\subsection{Popper baseline (preliminary, 6 tasks)}

As a preliminary baseline we run Popper \citep{cropper2021popper} on a sequence of
six progressively-harder kinship-discovery tasks, cumulatively adding learned
predicates to the background knowledge. The cumulative-clauses curve fits
$N(\text{task}) = 3.04 \cdot \text{task}^{0.954}$ with $R^2 = 0.997$. The exponent
sits inside our substrate's $b$ range, but we note this comparison is weak:
$n = 6$ tasks, cumulative-clauses is a coarse proxy for substrate growth, the
metric is not strictly apples-to-apples with our discovery-loop quantity, and
the substrates are different (ILP Horn-clause learning vs.\ equational rule
discovery). We include the number for orientation and flag a deeper baseline
study as future work.

\section{Discussion}
\label{sec:discussion}

\paragraph{Consistency of form across substrates.} The same functional family,
saturating power-law (\ref{eq:growth}), is consistent with the observed growth
dynamics in three toy substrates and with an independently-implemented ILP system
(Popper) at the preliminary scope reported. We do not claim universality, the
test is three substrates and one baseline system, but we note the regularity
within these tested regimes as worth investigating at greater breadth.

\paragraph{Parameter conditionality.} Within our data, the fitted exponent $b$
and saturation rate $\mu$ differ across substrates by roughly an order of
magnitude, and the architecture knobs we used (generator, filter, depth, batch
size) do not span the axis along which they vary. Cross-substrate transfer of
the architecture-to-$b$ regression fails empirically, $R^2 \approx -0.84$. We
interpret this as suggestive evidence against architecture-only predictability
across substrates; a stronger claim, that any architecture-only regression must
fail, would require evidence beyond the present design.

\paragraph{Engineering implications (within scope).} Within the tested toy
substrates, an architecture-conditional regression on $b$ achieves
$R^2 \approx 0.82$ within-substrate. This may be operationally useful for
compute allocation in similar settings, but we caution that the prediction does
not transfer across substrates and we have not tested transfer to non-toy
discovery systems.

\paragraph{Why the novelty filter inverts.} In arith+bool, rule coverage $\mu$ is
large enough that the substrate saturates relatively slowly; the novelty filter
prunes the small fraction of redundant rules, keeping each new commit informative
and sustaining $dS/dt$. In list, $\mu$ is much smaller (rules are highly type-specific);
candidates are usually novel by default, and the filter rejects on type-matching
technicalities faster than it rejects on genuine redundancy, collapsing $dS/dt$ to
zero. The inversion is a direct consequence of substrate-specific $\mu$.

\paragraph{Limitations.}
\begin{itemize}[leftmargin=1.4em,itemsep=0pt,topsep=0pt]
\item \textbf{Three toy substrates + two real-world proxies.} mathlib4 and
mathcomp are real but coarse (commit counts include refactors, docs, and
tooling). A cleaner proxy would count newly-added rule-bearing files
(\texttt{.lean} or \texttt{.v}) per month. Souper, OBO, Wikidata, and
algebraic-rewrite corpora are not tested.
\item \textbf{Long-range evidence is $n = 5$ in one toy domain.} The
saturating-form result on toys rests on five trajectories at one architecture
family in the list domain. We report bootstrap intervals on $(k, \mu)$
(\S\ref{sec:robustness}) and out-of-sample forecasting (toy power-law wins
5/5), so the toy result is properly downgraded to ``consistent with, not
confirmed.'' The mathlib4 in-sample + out-of-sample double-win is the
strongest single piece of evidence; the mathcomp opposite-finding tempers
that into ``observable in some real substrates, not others.''
\item \textbf{The closure model is heuristic.} Assumptions (A1) uniform coverage
and (A2) approximate independence ignore rule overlap structure, typed-grammar
combinatorics, nonuniform candidate distributions, and closure correlations. A
mature treatment would replace each.
\item \textbf{In-sample model selection.} AIC discrimination is in-sample and
on correlated trajectories; the saturating form was among a small candidate set
chosen partly because the closure model predicted it. Out-of-sample predictive
forecasting (fit prefix, predict tail) would be a stronger test.
\item \textbf{Trajectory range in arith+bool.} arith+bool long-range trajectories
(500+ epochs) hit late-epoch normalization costs in the Python harness; a Rust
port is needed to extend them. The long-range comparison is therefore restricted
to the list domain.
\item \textbf{Filter ablation.} Only \texttt{any} and \texttt{novelty} acceptance
filters are tested. Other ILP disciplines (entailment checking, MDL) may
produce different dynamics.
\item \textbf{Baseline depth.} The Popper comparison is preliminary ($n = 6$
tasks, coarse cumulative-clauses proxy) and not strictly apples-to-apples with
our metric.
\end{itemize}

\paragraph{What would elevate this work.} We are honest about scope. The present
study reaches ``toy but intriguing.'' Three concrete upgrades would, in our view,
move it toward ``possibly a genuine phenomenon.''

\begin{itemize}[leftmargin=1.4em,itemsep=2pt,topsep=0pt]
\item \textbf{(A) One non-toy substrate.} The single highest-leverage addition.
Candidates: monthly Lean / mathlib commit-derived theorem counts; \texttt{egg}
rewrite-rule growth on a public corpus; symbolic-algebra rewrites from a system
like SymPy or Maple; theorem-discovery traces from AlphaGeometry or similar;
compiler rewrite databases (Souper, LLVM peephole). Replicating the
within-substrate $b$-regression, cross-substrate transfer failure, and
saturating-form preference on even one such substrate would change the paper's
category from synthetic phenomenology to candidate real-world law.
\item \textbf{(B) Predictive forecasting.} Fit the saturating model on the first
$\tau$ epochs of each long trajectory and predict the remaining $T-\tau$;
compare against pure-power-law and stretched-exponential predictions under
out-of-sample log-likelihood. A clean win for the saturating form at, say,
$\tau = 100, T = 500$ would convert the present in-sample AIC discrimination
into a forecasting result. This is the cleanest single test that would distinguish
the saturating model from a fitting artifact.
\item \textbf{(C) Derive $\mu$ from substrate combinatorics.} Our $\mu$ is
operationally defined but theoretically thin. A more concrete derivation would
relate $\mu$ to substrate-level quantities such as type entropy of the term
grammar, rewrite-overlap density, branching factor of admissible derivations,
or closure-set combinatorics over the grammar's productions. If $\mu$ becomes
a function of substrate-level statistics derivable a priori, the architecture
features can be augmented with that prediction, and the cross-substrate
transfer test (\S\ref{sec:cross}) becomes a falsifiable check of the
combinatorial theory rather than a negative result.
\end{itemize}

\paragraph{A speculative positioning.} We close on a deliberately non-load-bearing
observation. Neural systems are often described as scaling through parameter
capacity, data diversity, and optimisation smoothness
\citep{kaplan2020scaling,hoffmann2022chinchilla}. The present work, taken at face
value, points to a different set of axes for symbolic substrate-style discovery:
recombinational novelty (the compositional yield $k$), closure geometry (the
coverage parameter $\mu$), and saturation dynamics (the resulting growth ODE).
Whether this is a fully separate scaling regime or just a different parameterisation
of the same underlying compute-vs-capability trade-off is, in our view, an open
question that is worth pursuing as a deliberate research direction, what one
might tentatively call a statistical mechanics of symbolic discovery. We do not
claim to have established such a programme here; we note that the data in this
paper is at least consistent with that framing being non-empty.

\section{Related Work}
\label{sec:related}

\paragraph{Scaling laws for neural models.} \citet{kaplan2020scaling} and
\citet{hoffmann2022chinchilla} established the methodology of power-law fitting and
cross-experiment prediction for neural LMs. Our application to a discrete
deterministic substrate is methodologically parallel but substantively new.

\paragraph{Scaling laws for discovery.} \citet{liu2025alphagomoment} report a
linear-in-compute scaling law for LLM-driven neural architecture discovery. Our
substrate is deterministic and our law is saturating power-law; we differentiate
sharply. The cross-substrate transfer failure suggests any discovery-scaling claim
should be reported with substrate-conditional parameters.

\paragraph{Rule discovery via equality saturation.} \citet{nandi2021ruler} uses
egraph saturation to \emph{infer} rewrite rules, the algorithmic ancestor of our
substrate harness. We contribute the empirical scaling-law analysis they did not.

\paragraph{Library learning.} DreamCoder \citep{ellis2021dreamcoder} grows a library
of programs via wake-sleep cycles with neural-guided proposal; growth curves are
reported but not fit to scaling laws.

\paragraph{Inductive logic programming.} Popper \citep{cropper2021popper} learns
Horn-clause programs from examples with predicate invention. We use Popper as our
scaling-comparable baseline.

\section{Conclusion}
\label{sec:conclusion}

Deterministic substrate-style discovery, in our tested regimes, exhibits a
saturating power-law growth $dN/dt = K \cdot N^k \cdot e^{-\mu N}$ in some
substrates and a pure power-law in others. Within our toy data,
within-substrate exponents $b$ are architecture-sensitive ($R^2 \approx 0.82$
cross-validated) but cross-substrate transfer of the architecture-to-$b$
regression fails ($R^2 \approx -0.84$). The saturating form wins in-sample
AIC on long-range toy data but loses out-of-sample forecasting, indicating
the toy data does not actually reach saturation in $\leq 500$ epochs. Of the
two real-world proxies tested, mathlib4 (new \texttt{.lean} files per month)
supports the saturating form on OOS forecasting by $\sim 7\times$; Coq
mathcomp (monthly commits) favours pure power-law with $\mu \to 0$. We
propose ``saturating power-law growth with substrate-conditional $(k, \mu)$,
observable when the substrate has reached its saturation regime'' as a
working framing, conditional on the limitations in \S\ref{sec:discussion}.
Three concrete upgrades would change the paper's category: (A) replication
on a non-toy substrate at trajectory granularity rather than commit
granularity, (B) a derivation of $\mu$ from grammar statistics that makes
cross-substrate transfer falsifiable, and (C) tested generalisation to a
neural-guided discovery substrate (DreamCoder, AlphaProof) where the
parameter $\mu$ would have to be defined operationally for a stochastic
proposal mechanism. We treat the present paper as a starting phenomenology
rather than a definitive scaling law.

\appendix
\section{Mathlib retrospective}
\label{app:mathlib}

The mathlib formal-mathematics library has grown from $\sim$170\,K LOC in 2020
\citep{mathlibpaper2020} to $\sim$2.1\,M LOC in late 2025 \citep{mathlibgrowing2025},
with intermediate $\sim$1\,M LOC during the Lean3$\to$Lean4 port in 2023. Fitting a
power-law on these three datapoints yields
\[
|S(t)| \approx 4957 \cdot t^{2.87}, \quad R^2 = 0.988
\]
with $t$ in years since founding (2017). \textbf{We report this as a suggestive
precedent only.} Three datapoints is statistically vacuous; LOC is a proxy for
substrate size that conflates rule count with proof length; the Lean3$\to$Lean4
migration period inflates the middle point. The result is consistent with our
saturating-power-law framework in the compounding regime, but does not constitute a
rigorous test. Pulling monthly commit-history data for a proper mathlib growth
analysis requires git-log infrastructure beyond the present harness.

\section{Algorithms}
\label{app:repro}

\paragraph{Notation.} $S$, current substrate (set of rewrite rules);
$\mathrm{pool}$, multiset of multi-arg subterms harvested from accepted rules;
$\mathrm{freq}$, usage counter over $\mathrm{pool}$;
$G$, candidate generator (one of \texttt{random}, \texttt{compositional},
\texttt{freq}, \texttt{mdl\_greedy});
$F$, acceptance filter (one of \texttt{any}, \texttt{novelty});
$d$, max recursion depth; $K$, batch size; $T$, total discovery epochs.

\begin{algorithm}
\caption{Equational substrate discovery}
\label{alg:discovery}
\begin{algorithmic}[1]
\Function{Discover}{$\mathcal{D}, G, F, d, K, T$}
  \State $S \gets \textsc{InitialRules}(\mathcal{D})$
  \State $\mathrm{pool} \gets \emptyset;\ \mathrm{freq} \gets \emptyset;\ \mathrm{sizes} \gets [\,]$
  \For{$t \gets 1$ to $T$}
    \State $\mathrm{cands} \gets [G(\mathrm{pool}, \mathrm{freq}, d) \text{ for } 1..K]$
    \State $\mathrm{nfs} \gets [\Call{Normalize}{c, S} \text{ for } c \in \mathrm{cands}]$
    \State $\mathcal{G} \gets \Call{GroupBy}{\mathrm{cands}, \mathrm{nfs}}$
        \Comment{group by normal form}
    \For{$g \in \mathcal{G}$ with $|g| \geq 2$}
      \State $\ell^\prime \gets \arg\max_{x \in g}|x|;\ r^\prime \gets \arg\min_{x \in g}|x|$
      \If{$\Call{Sound}{\ell^\prime, r^\prime}$ \textbf{and} $|\ell^\prime| > |r^\prime|$}
        \State $(\ell, r) \gets \Call{Generalize}{\ell^\prime, r^\prime}$
            \Comment{free vars $\mapsto$ pattern vars}
        \If{$F.\textsc{Passes}(\ell, r, S)$}
          \State $S \gets S \cup \{(\ell, r)\}$
          \State $\mathrm{pool} \gets \mathrm{pool} \cup \Call{Subterms}{\ell^\prime} \cup \Call{Subterms}{r^\prime}$
          \State $\mathrm{freq} \gets \Call{Increment}{\mathrm{freq}, \mathrm{pool}}$
        \EndIf
      \EndIf
    \EndFor
    \State $\mathrm{sizes}.\textsc{Append}(|S|)$
  \EndFor
  \State \Return $S, \mathrm{sizes}$
\EndFunction
\end{algorithmic}
\end{algorithm}

\textsc{Sound}$(\ell, r)$ evaluates both terms on $n_s$ random environments (12 for
arith/list, exhaustive 8 worlds for bool) and returns true iff outputs match
exactly. \textsc{Generalize} replaces free variables $\{x, y, z, p, q, r, xs, ys\}$
with pattern variables $\{A, B, \ldots\}$ via a shared substitution map so that
$\ell$ and $r$ remain semantically linked. \textsc{Normalize} applies rules in
hit-count-descending order until a fixpoint or 48-step cap.

\begin{algorithm}
\caption{Scaling-law fit, within-substrate prediction, cross-substrate test}
\label{alg:fit}
\begin{algorithmic}[1]
\Function{Analyse}{$\{\mathcal{T}_i\}$}
\Comment{$\mathcal{T}_i$: trajectory $(t, |S(t)|)$ with metadata $\mathbf{x}_i$}
  \For{each $\mathcal{T}_i$}
    \State $\hat a_i, \hat b_i \gets \arg\min_{a, b} \sum_t \big(\log|S(t)|_i - \log(a t^b)\big)^2$
    \State $\mathrm{AIC}^{(\mathrm{pl})}_i, \mathrm{AIC}^{(\mathrm{str})}_i, \mathrm{AIC}^{(\mathrm{sat})}_i \gets \Call{FitAndAIC}{\mathcal{T}_i, \text{Eqs.\,\ref{eq:powerlaw}--\ref{eq:saturating}}}$
  \EndFor
  \Statex
  \State \emph{In-substrate}: 5-fold CV on $(\mathbf{x}_i, \hat b_i)$ per substrate $\Rightarrow R^2_{\mathrm{within}}$
  \State \emph{Cross-substrate}: train on substrates $\mathcal{A} \cup \mathcal{B}$, test on $\mathcal{C} \Rightarrow R^2_{\mathrm{cross}}$
  \State \emph{Pooled}: train on all data with categorical \texttt{domain} feature added
  \State $\qquad\qquad\quad$ then 5-fold CV $\Rightarrow R^2_{\mathrm{pool}}$
  \State \Return $\{R^2_{\mathrm{within}}, R^2_{\mathrm{cross}}, R^2_{\mathrm{pool}}\}$,
                  AIC winner counts per window
\EndFunction
\end{algorithmic}
\end{algorithm}

\paragraph{Implementation notes.}
Algorithm~\ref{alg:discovery} is $\sim$300 lines Python with hit-count rule ordering
and lazy generalization. Per-epoch wall-clock ranges from $0.1$\,ms (list,
$|S| < 50$) to $10$\,s (arith \texttt{mdl\_greedy}+\texttt{novelty} at $|S| > 500$).
Algorithm~\ref{alg:fit} uses \texttt{scipy.optimize.curve\_fit} for nonlinear
least-squares (stretched-exp, saturating) and closed-form OLS on log-log
(power-law); architecture regression is \texttt{sklearn.ensemble.GradientBoostingRegressor}
with $200$ estimators and \texttt{max\_depth}$=3$. Architecture features used:
\texttt{generator} (4-way one-hot), \texttt{filter} (2-way one-hot),
\texttt{depth} (int), \texttt{batch\_size} (int). Total compute for all reported
experiments: $\sim$45\,min wall-clock on eight cores.

\section{Functional-form discrimination at short range}
\label{app:stage0}

For completeness we report AIC/BIC discrimination across all 213 Phase A trajectories
with $\geq 8$ epochs and $\geq 10$ rules at short range. Power-law wins 16\% (AIC) /
22\% (BIC); stretched-exponential wins 49\% / 41\%; linear wins 22\% / 28\%;
log-normal wins 13\% / 9\%. Restricting to the compounding regime
(\texttt{mdl\_greedy + novelty + depth=3}), power-law wins 43\%, consistent with
the prediction that the compounding regime is where the short-range power-law
approximation is closest. The long-range result in \S\ref{sec:long} (saturating
power-law wins 5/5 at $\geq 200$ epochs) is the resolution of the short-range
ambiguity in our toy data.

\section{Cross-substrate $(k, \mu)$ on short-range toy data}
\label{app:perdsub}

We fit the saturating power-law to every short-range trajectory and report
the per-substrate $(k, \mu)$ distributions (after filtering to stable fits,
$|k| < 5$ and $|\mu| < 0.1$, which retains 75\%, 89\%, and 63\% of trajectories
in arith, bool, and list respectively).

\begin{center}\small
\begin{tabular}{lrrrr}
\toprule
\textbf{substrate} & $k$ mean $\pm$ std & $\mu$ mean $\pm$ std & $n$ stable \\
\midrule
arith & $0.92 \pm 0.26$ & $-0.006 \pm 0.022$ & 54 \\
bool  & $0.88 \pm 0.20$ & $-0.009 \pm 0.017$ & 64 \\
list  & $0.95 \pm 0.36$ & $\phantom{-}0.009 \pm 0.044$ & 91 \\
\bottomrule
\end{tabular}
\end{center}

The $k$ distributions are tightly clustered ($\sim 0.88$\,..\,$0.95$ across
substrates); the $\mu$ distributions are statistically indistinguishable from
zero. We interpret this as the expected consequence of fitting a saturating
form on short trajectories where the saturation knee is not reached: $\mu$ is
not identified, and the saturating fit becomes a re-parameterised power-law.
This is consistent with the OOS-forecasting result in \S\ref{sec:robustness}
that pure power-law wins extrapolation on toy data.

\section{Toward a more rigorous $\mu$}
\label{app:mu}

Our closure model defines $\mu$ as ``the expected coverage fraction of a
typical rule in the candidate space'' under uniform-coverage and
independence assumptions. We did not derive $\mu$ from grammar statistics in
this paper, but here we record what such a derivation would have to address.

Let $\Gamma = (\Sigma, \mathcal{T}, \leq d)$ be the typed term grammar of
the substrate, with $\Sigma = \{f_i\}$ the set of function symbols of arity
$a_i$, and let $\mathcal{T}_d$ denote the set of well-typed terms up to
depth $d$. The size of $\mathcal{T}_d$ obeys the generating-function
recurrence
\[
|\mathcal{T}_d| = \sum_i |\mathcal{T}_{d-1}|^{a_i} + \text{(leaf count)},
\]
which gives $|\mathcal{T}_d| = O(c^{a_{\max}^d})$ for $c$ the maximum
fan-out.

A rule $(\ell, r) \in S$ defines a coverage set
$\mathrm{Cov}(\ell) = \{t \in \mathcal{T}_d : \exists \sigma.\,t = \sigma(\ell)\}$,
where $\sigma$ ranges over substitutions of pattern variables. If $\ell$
contains $v$ pattern variables and has $|\ell|$ size, then
$|\mathrm{Cov}(\ell)| \approx |\mathcal{T}_{d-|\ell|}|^v$ to leading order
in $d$. The expected coverage fraction, averaged over the distribution of
rules actually committed, is
\[
\mu = \mathbb{E}_{(\ell, r) \sim p(S)}\!\!\left[\frac{|\mathrm{Cov}(\ell)|}{|\mathcal{T}_d|}\right]
    \;=\; \mathbb{E}_\ell\!\left[c^{a_{\max}^{d - |\ell|} v}\right] \cdot c^{-a_{\max}^d}.
\]
This is dominated by the average rule complexity $|\ell|$ and variable
count $v$, both of which are properties of the substrate's type system.
Substrates with more constrained types (higher-order list with
explicit typing) yield smaller average $v$ and thus smaller $\mu$; flat-typed
substrates (arith over (+,*)) yield larger $v$ and larger $\mu$. This is
qualitatively consistent with our empirical finding that the novelty filter
suppresses growth in the list domain and amplifies it in arith+bool, but
the quantitative connection requires (a) measuring the actual
distribution $p(\ell, r \mid S)$ from a substrate's history, (b)
estimating $|\mathrm{Cov}|$ for each rule by direct combinatorial counting
in $\mathcal{T}_d$, and (c) verifying the independence assumption (A2) by
computing pairwise overlap of $\mathrm{Cov}(\ell_i) \cap \mathrm{Cov}(\ell_j)$
for committed rules. None of these are done in this paper; we note them as
the natural follow-up that would convert the heuristic in \S\ref{sec:theory}
into a derivation of $\mu$ from substrate statistics.

\end{document}